\documentclass[runningheads]{llncs}
\usepackage{graphicx}
\usepackage{xcolor}
\usepackage{multirow}
\usepackage{float}

\usepackage{color}
\usepackage{colortbl}
\definecolor{LightGray}{rgb}{0.9,0.9,0.9}
\begin{document}
\title{CYRUS Soccer Simulation 2D Team Description Paper 2022}
%
%\titlerunning{Abbreviated paper title}
% If the paper title is too long for the running head, you can set
% an abbreviated paper title here
%
\author{
Nader Zare\inst{1}\and 
Arad Firouzkouhi\inst{4}\and
Omid Amini\inst{5}\and
Mahtab Sarvmaili\inst{1}\and
Aref Sayareh\inst{6}\and
Saba Ramezani Rad\inst{4}\and
Stan Matwin\inst{1}\inst{2} \and
Amilcar Soares\inst{3}}
\authorrunning{N. Zare et al.}
% First names are abbreviated in the running head.
% If there are more than two authors, 'et al.' is used.
%
\institute{
Institute for Big Data Analytics, Dalhousie University, Halifax, Canada\\
\and
Institute for Computer Science, Polish Academy of Sciences, Warsaw, Poland\\
\and
Memorial University of Newfoundland, St. John's, Canada\\
\and
Amirkabir University of Technology, Iran\\
\and
Qom University of Technology, Iran\\
\and
Shiraz University, Iran\\
\email{\{nader.zare, mahtab.sarvmaili\}@dal.ca},\\ \email{\{arad.firouzkouhi, saba\_ramezani\}@aut.ac.ir}\\ 
\email{\{arefsayareh, omidamini360\}@gmail.com},\\
\email{stan@cs.dal.ca}, \email{amilcarsj@mun.ca}
}
\maketitle              % typeset the header of the contribution
\begin{abstract}
Soccer Simulation 2D League is one of the major leagues of RoboCup competitions. In a Soccer Simulation 2D (SS2D) game, two teams of 11 players and one coach compete against each other. The players are only allowed to communicate with the server that is called Soccer Simulation Server. This paper introduces the previous and current research of the CYRUS soccer simulation team, the champion of RoboCup 2021. We will present our idea about improving Unmarking Decisioning and Positioning by using Pass Prediction Deep Neural Network. Based on our experimental results, this idea proven to be effective on increasing the winning rate of Cyrus against opponents.

\keywords{Pass Prediction  \and Deep Neural Network \and Soccer Simulation.}
\end{abstract}
\section{Introduction}
RoboCup has been holding robotic soccer competitions annually since 1997. The idea of robotic soccer games had been proposed as a novel research topic in 1992. The main goal of this competition is implementation of robotic soccer games as a research area to improve the robotics, therefore it consists of several leagues such as rescue, soccer simulation, and standard platform soccer to advance every aspect of Robotics. 

Soccer simulation 2D (SS2D) league is one of the oldest RoboCup leagues. CYRUS team has been participating in RoboCup competitions since 2013 and won the championship of SS2D league in RoboCup 2021\cite{cyruschampion}. This team has gained the first, second, third, fourth, and fifth places in RoboCup 2021, 2018, 2019, 2017, 2014, respectively. Also, CYRUS won first place in IranOpen 2021, 2018, and 2014, RoboCup Asia-Pacific 2018, and second place in JapanOpen 2020. CYRUS uses Agent2D as its base code \cite{agent2d}. 
In this paper, we briefly explain a novel idea about Unmarking and Positioning using Pass Prediction Neural Network. 

\vspace{-0.5cm}
\subsection{Related Work}
In recent years, SS2D teams improved their performance by implementing novel ideas, such as Player's MatchUp. Helios has developed an algorithm called Player’s MatchUp for exchanging players’ positions during a game \cite{hel21}. They took advantage of the distributed representation of actions and players to find similarities among teams' behavior \cite{hel19}. FRA-United applied reinforcement learning algorithm in the context of SS2D to improve marking behavior \cite{fra19,fra21}. Oxsy used the RSA accumulator(Real-World Performance of Cryptographic Accumulators) to model opponents strategy by using the observation coach. Oxsy members have implemented an idea called adaptive offside trap\cite{oxsy18,oxsy21}. Razi improved their offensive performance by updating chain action evaluator function and gravity strategy\cite{razi18,razi19}. Persepolis has implemented a Path Planning algorithm to improve its offensive strategy\cite{pers21}. Hades improved their marking algorithm by evaluating the opponent player's position\cite{had21}. 
\vspace{-0.5cm}
\subsection{Previous Work}
CYRUS was established in 2013 and participated in many competitions since then. In this section, we briefly explain about the history of our research . We have stared working on the passing behavior of team, unmarking strategy and team's formation\cite{cyrus13,cyrus14}. 
Then after, we focused on the defensive behavior \cite{cyrus15}, shooting algorithms and communication \cite{cyrus16} of our team.
In 2018 we developed an opponent behavior prediction module \cite{cyrus18}.
Since 2018, we have mainly focused on the development of Pass prediction module and Data Generator framework which are elucidated in \cite{cyrus19,cyrus21,cyrussamposiom}.
In recent years, we have release some parts of our research such as CYRUS 2014 Source Code, PYRUS - Python SS2D Base Code, SS2D Log Analyzer.
\vspace{-0.3cm}
\subsubsection{CYRUS 2014 Source Code}
One of the major contribution of Cyrus to SS2D community was publicly releasing the Cyrus 2014 source code. The code is available on our GitHub\footnote{Cyrus 2014 Source \url{https://github.com/naderzare/cyrus2014}.}.
\vspace{-0.3cm}
\subsubsection{PYRUS - Python Soccer Simulation 2D Base Code}
Due to the high performance of C++ , majority of SS2D teams' base code is written in C++ such as Helios Base\cite{agent2d}, Gliders Base\cite{glbase}, WrightEagle Base\cite{wr}. Despite  its processing time, the implementation of novel AI and machine learning algorithms are very much complicated, and requires lots of coding and debugging. This would be a tedious and time consuming task for the teams.
Python on the other hand is simple and very well supported in terms of AI and machine learning libraries such as Scikit learn \cite{scikit}, PyTorch \cite{pytorch} and TensorFlow \cite{tensor}.
Therefore we have decided to develop a Python 2D base accessible for all of the SS2D community.
This base is available in Cyrus Github\footnote{Pyrus Base Code Source \url{git@github.com:Cyrus2D/Pyrus2D.git}.}, covering all features in the Agent2D when server runs on FULL-STATE mode.

\section{Pass Prediction}
As in real soccer, passing is one of the possible actions that can lead the ball to the goal. Predicting the pass target player, from the point of view of the ball owner, has many benefits in defensive and offensive algorithms. For example, opponent's pass prediction can enhance marking decision algorithm, and teammate's pass prediction can increase the efficiency of the unmarking strategy.
In this paper, the ball owner is the player who can kick the ball in the current cycle or receive the ball in the future cycles.
To predict the behavior of (our) ball owner, we were required to create the dataset of game states from this player point of view. For this purpose, we embedded a Data Extractor module in each one of the players and then we recorded the features of game states and their corresponding label. The label shows the uniform number of the player who is target of the best pass.

\begin{figure}
    \centering
    \includegraphics[scale=0.20]{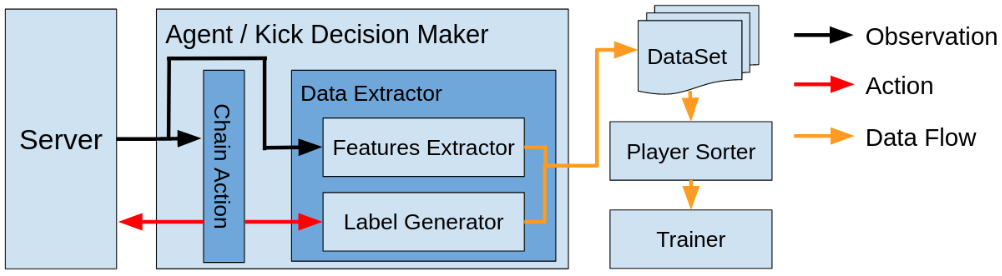} \\
    \caption{Overview of the Data Extractor module.}
    \label{fig:dataflow}
\end{figure}

This module contains two sub-modules, Features Extractor and Label Generator (Figure \ref{fig:dataflow}).
The Features Extractor sub-module perceives the state of the game, and generates features of the ball, opponent players, and our players by using the observation of the ball owner. The Label Generator module uses the output of the Chain-Action module to assign the corresponding action related to the features as a label by the Features Extractor sub-module.
After generating the data set, the Player Sorter module sorts the features by using x of players' positions for opponents and the uniform numbers of players for our players.
The structure of the Data Extractor module is presented in Figure \ref{fig:dataflow}, and more details are available in our recent papers\cite{cyrus21,cyruschampion,cyrussamposiom}.

\vspace{-1cm}
\section{Unmarking}
Unmarking is an offensive action that players of the offender team use to escape from opponent marking and find a good spot for receiving a pass (when they are not the ball owner). The player who wants to unmark is called Unmarker.
Unmarking includes two main parts, "Decisioning" and "Positioning".
"Decisioning" finds the pass sender, so in this part, the player discovers one teammate that will send a pass to itself in the future. The positioning parts look for the best spot to receive the future pass.

\subsection{Unmark Decisioning by using Pass Prediction module}
When one of our players posses the ball, the rest of players should assist it for obtaining the best chain action to move the ball forward. For this purpose, each player of CYRUS tries to simulate a tree that contains the passes and game states generated in the future. This tree includes the probable passes and their outcome states in the future. Each node of the tree contains a state of the game so that one of our players is the ball owner in the state. Each edge of the tree shows a probable pass in the future.

\textbf{Pass Predictor} To create the tree, we developed a module called Pass Predictor to predict the target player of the pass in a given state. This module receives a state and a list of players called "Ignored Players". The Ignored Players presents players who can not be the pass receiver in the given state. To find the best passes, the Pass Predictor module generates features from the given state and feeds them to the trained pass prediction DNN and then produces the probability of receiving the pass for all players. There after, it finds the players who have the maximum probability among players that are not in the "Ignored Players" list. Finally, it creates an object called Pass-State-Value containing pass, its outcome state, and state value. 
%To estimate the outcome state of a pass, we change the position of the ball to the position of the target player. 
The state value stands to show the probability of the pass. Eventually, this module returns two passes who can be the pass receiver in the state, their result state, and state point.  

\textbf{Ignored Player list} The Ignored Player list contains players who are ball owners in nodes of the tree. Therefore, a player can be the ball owner in one node of the tree.

\textbf{Root node} The root node of the tree is created by using the current state of the game if one of our players can kick the ball right now. However, if one of our player receives the ball in the next few steps, the unmarker predicts the cycle of intercepting the ball and the position of ball receiver, then it updates the root node of pass predictor tree.

% next steps 
% \textbf{Candidates List} We store the pass-state-value in a list a.k.a "Candidates". In the algorithm, we select the object with higher value among the objects and feed it to Pass Predictor module, then we add outputs of the Pass Predictor module into the list.

% To initial the algorithm, the unmarker inserts 
After creating the root node, the unmarker player feeds the state of the root node and the Ignored Players list to the Pass Predictor module and then it receives the list of two Pass-State-Value objects. 
At this step, the Ignored Players will be updated to include the ball owner. After receiving the Pass-State-Value objects, the unmarker inserts them into a list called Candidate List. 
%The unmarker initially staarts with the current ball owner and then it expands the tree using the obtained Pass-State-Values and Best First Search. It chooses 
To grow the tree, we use the Best First Search algorithm in which, at the next step, the unmarker selects the best object from the Candidate List with highest value, and then it added the selected object to the tree. Also, the ball owner of the state is added to the Ignored Players list. This procedure continues until the number of tree nodes is equal to ten. 
After termination of this procedure, the umarker agent looks for its corresponding node in the tree, then it chooses the parent player as the "Passer" for the unmarking procedure in order to receive the pass from it in the future.
% After creating the tree, the unmarker finds the node that its ball owner is itself in the node's state. The ball owner in the parent node of its node is the player that unmarker should 

\begin{figure}
    \centering
    \includegraphics[scale=0.12]{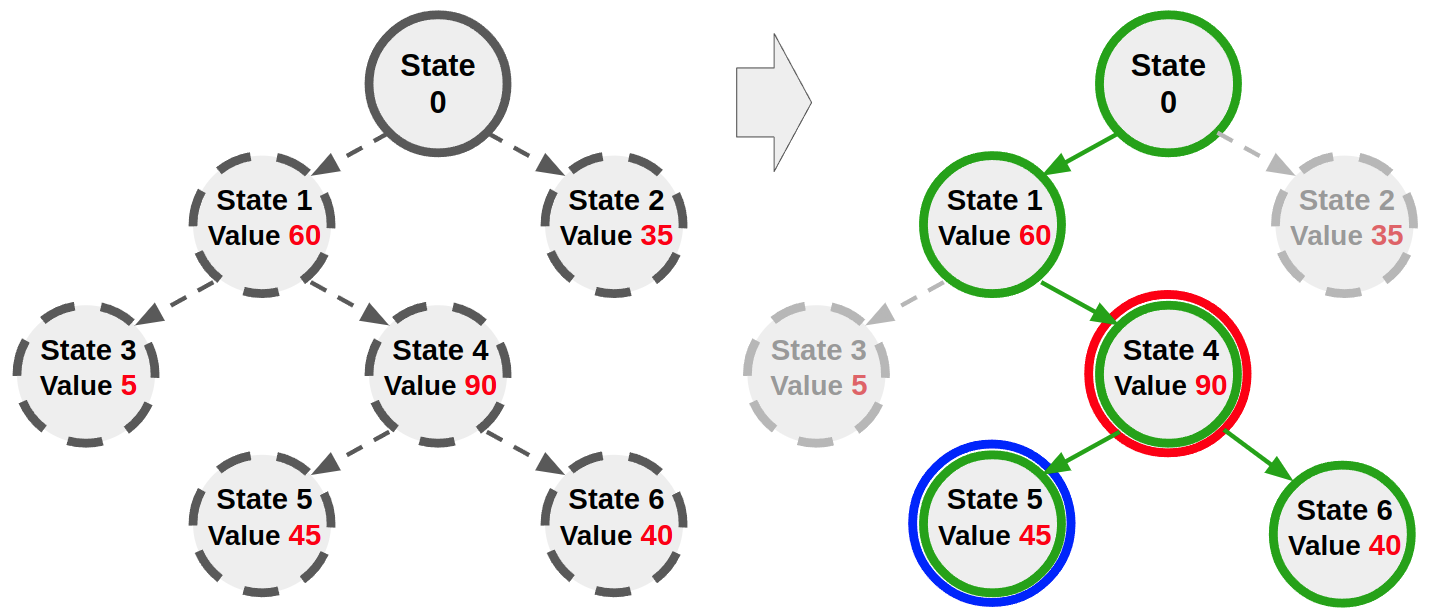} \\
    \caption{Overview of the Unmark Decisioning algorithm. The left tree shows the result states and their points. The green circles in the right tree presents the nodes have been selected as best node from the Candidate list. The blue circle shows the node that the unmarker is ball owner in its state, and its parent node is shown by red circle.}
    \label{fig:unmarking}
\end{figure}

Figure \ref{fig:unmarking} shows an overview of the Unmarking Decisioning.

\vspace{-0.5cm}
\subsection{Positioning in unmark}
After finding the "Passer" player, the unmarker should find a position to receive the pass. For this purpose, the unmarker generates a set of points around its current position in three levels that respectively have 2, 4, 8 meters as radius. Each level contains 8 points.
Selecting the best position contains two steps. First, the unmarker simulates a pass from the Passer Player to each position, if the unmarker can receive the ball in the position before the opponent players the position is a validated point.
Then, among the valid points, the unmarker selects the closest position to the opponent's and it tries to move to the best position in the future cycles.

\begin{figure}[h!]
    \centering
    \includegraphics[scale=0.20]{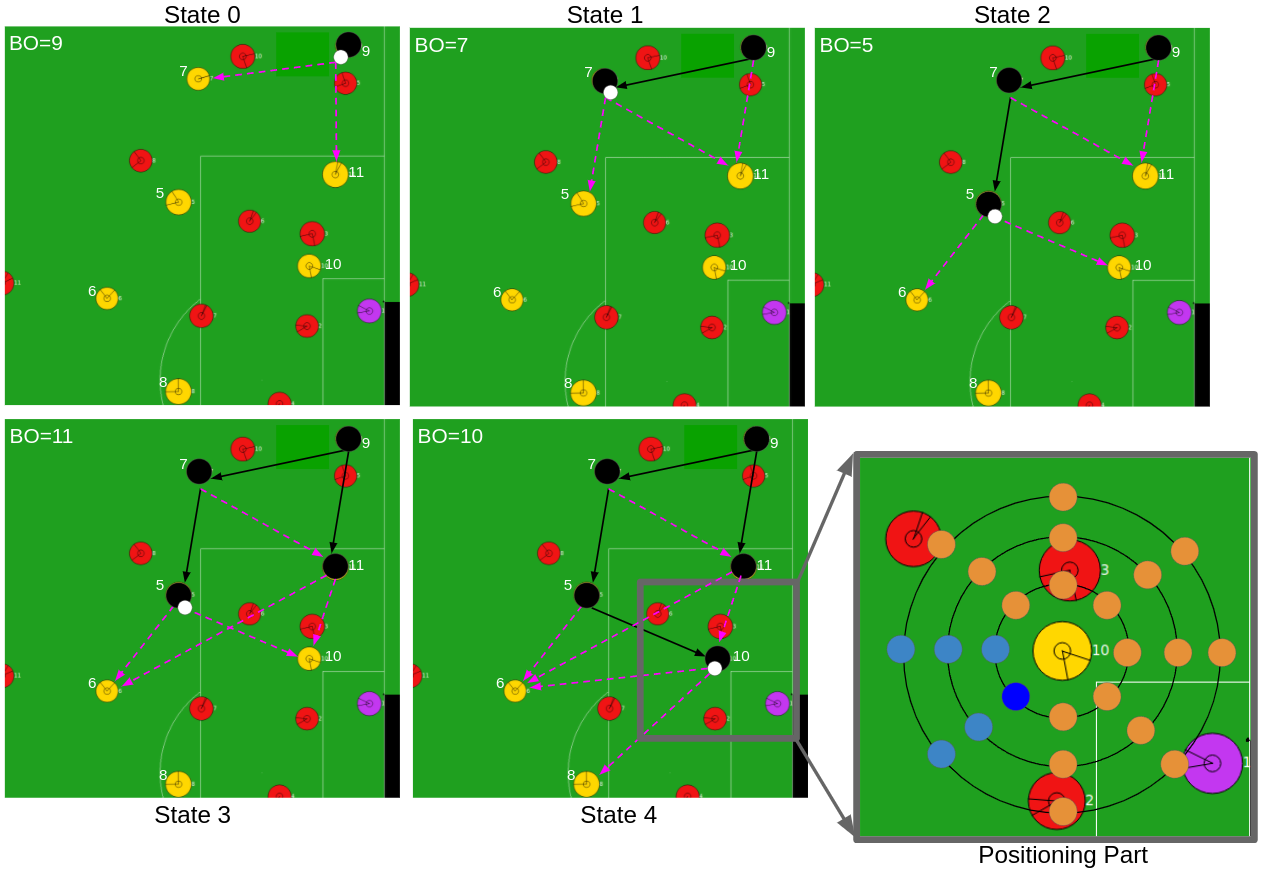} \\
    \caption{Overview of the Unmarking algorithm. Player 10 would like to unmark, and Player 9 is the current Ball Owner in the current cycle of the game. States show the future state of the game. Pink dashed arrows present candidate passes, and black arrows show the selected best passes. Black circles present players in Ignored Player list. The best targets for Player 9 are Player 7 and Player 11 in state 0. Player 7 is selected to be the ball owner, and the best targets are Player 5 and Player 11 in the next step.  
    The pass from Player 7 to Player 5 is selected in state 2. In the next step, Player 11 is selected as the receiver, and in the state for Player 10 is the ball receiver from Player 5. The black arrows in State 4 show that the unmarker player should find a position to receive a pass from Player 5. The Positioning Part presents all position candidates for unmarker. Blue circles are validated positions and the dark blue circle is the best position for unmarking. Orange circles shows positions that the Player 10 can not receive a pass there.}
    \label{fig:unma}
\end{figure}

Figure \ref{fig:unma} shows an example of the Unmark Decisioning and Unmark Positioning parts in SS2D field.
\section{Results}
To obtain the data set, we ran 10,000 games between CYRUS and YuShan2021, and we obtained 3,000,000 data instances. We split data into two subsets, 80\% for training and 20\% for testing. The prediction model (DNN) has three layers of 350, 250, and 11 neurons, with RELU activation function and a softmax function at the last layer. %The activation function of internal layers is RELU; the last layer uses the SoftMax function. 
The validation accuracy of the trained neural network on the test data was 73\%.

Three unmarking algorithms are available in CYRUS, Pass Prediction Unmarking, Hard-Coded Unmarking, Voronoi Unmarking. The Pass Prediction Unmarking uses Deep Neural Network for decisioning, and searching around the player for positioning. Hard-Coded algorithm, the ball owner is passer if the ball owner is closer than 20 meters to the unmarker, otherwise the selected passer is a player that is close to the line between ball owner and the agent. In the Voronoi Unmarking, players uses Voronoi diagram to find the best position to escape from opponent's marking.
In CYRUS2021, we used Hard-Coded and Voronoi Unmarking algorithms, and if the first algorithm does not find a position, then the second algorithm tries to find a new position.

To test the benefits of Pass Prediction Unmarking algorithm, we created 6 versions of CYRUS. In the first version, we just used the Pass Prediction Unmarking. Pass Prediction and Voronoi Unmarking algorithms were used in the second version. Four other versions are presented in Table \ref{tab:my_label}. To test the benefits of the new algorithm, we ran 1000 games against YuShan2021 for each versions. All of the results are presented in Table \ref{tab:my_label}.
\begin{table}[]
    \centering
    \begin{tabular}{|l|l|l|l|l|l|l|l|l|l|l|l|l|}
    \hline
        Versions
         & \rotatebox{90}{Pass Prediction} 
         & \rotatebox{90}{Hard-Coded} 
         & \rotatebox{90}{Voronoi}	
         & \rotatebox{90}{Goals Scored}	
         & \rotatebox{90}{Goals Conceded}
         & \rotatebox{90}{Goal Difference} 
         & \rotatebox{90}{Win-rate}	
         & \rotatebox{90}{Expected-Win-rate}	
         & \rotatebox{90}{Pass Numbers} 
         & \rotatebox{90}{Pass Accuracy} 
         & \rotatebox{90}{Possession} 
         & \rotatebox{90}{Shoot Number} \\
         \hline
         \rowcolor{LightGray}
        V1 &Y&	&	&\textbf{1.13}&	0.70&	\textbf{0.43}&	45.75&	66.42&	313.54&	\textbf{0.90}&	62.18&	\textbf{3.28}\\
        V2 &Y&	& Y	&1.10&	\textbf{0.67}&	\textbf{0.43}&	\textbf{45.80}&	\textbf{67.55}&	\textbf{315.90}&	\textbf{0.90}&	\textbf{62.26}&	3.26\\
        \rowcolor{LightGray}
        V3 (CYRUS2021) & & Y& Y	&1.04&	0.78&	0.26&	39.80&	61.42&	307.35&	0.89&	61.36&	3.15\\
        V4 & & Y& 	&0.98&	0.70&	0.28&	41.70&	60.79&	308.30&	0.89&	61.44&	3.11\\
        \rowcolor{LightGray}
        V5 & &	& Y	&1.08&	0.76&	0.32&	43.40&	63.27&	287.69&	0.88&	59.86&	3.07\\
        V6 & &	& 	&0.92&	0.70&	0.22&	39.70&	59.79&	283.46&	0.88&	59.69&	2.85\\
        \hline
    \end{tabular}
    \caption{The impact of different unmarking approaches on Cyrus performance}
    \label{tab:my_label}
\end{table}

The obtained results proves the significance of our proposed method in the improvement of statistics against YuShan2021. For example, the new algorithm increased our expected win-rate for 5\%, also it improves our ball possession percentage and number of shoots.
\vspace{-0.5cm}
\section{Conclusion}
In this paper we proposed an novel idea for the unmarking strategy of CYRUS by Pass Predicting. 
To model our teammates’ passing behavior in the game, we exploited the generated features and trained a Deep Neural Network called Pass Predictor. To select a player as Passer for another agent, we used the Pass Predictor module to create a tree when the current state of the game is its root node. To search and grow the tree, we used the Best First Search algorithm. Finally, we explain an idea to find the best position for unmarking.

\section{Future Work}
We intend to explore other challenging aspects of SS2D and unmarking behavior in Soccer Simulation 2D and Unmarking. First, we would like to use the Pass Predictor module to find the best position for catching  a pass by relocating the position of unmarker and feeding it to the DNN. Also, we are willing to create a model with more generality on the wider range of teams.
%
% ---- Bibliography ----
%
% BibTeX users should specify bibliography style 'splncs04'.
% References will then be sorted and formatted in the correct style.
%
% \bibliographystyle{splncs04}
% \bibliography{mybibliography}
%

\end{document}